\documentclass[11pt]{article}

\usepackage[preprint]{acl}

\usepackage{times}
\usepackage{latexsym}
\usepackage[T1]{fontenc}
\usepackage[utf8]{inputenc}
\usepackage{microtype}
\usepackage{amsmath}
\usepackage{amssymb}
\usepackage{booktabs}
\usepackage{graphicx}
\usepackage{xcolor}
\usepackage{enumitem}

\newcommand{\bench}{AstronautBench}
\newcommand{\qa}{Entity Knowledge QA}
\newcommand{\gpt}{GPT-6.1 Sol}
\newcommand{\gem}{Gemini 4 Argon}

\title{\bench: Measuring Agent Self-Improvement Across Episodes\\Through Knowledge, Evidence, and Progress}
\author{Congyuan Yu \\
  Carnegie Mellon University \\
  \texttt{congyuay@andrew.cmu.edu} \And
  Runping Tian \\
  \texttt{tianrunping2026@outlook.com}}

\begin{document}
\maketitle

\begin{abstract}
LLM agents are expected to improve themselves from their own experience across episodes, with fixed weights. Self-improvement in an embodied world cannot be read from one measure. Existing benchmarks measure task progress, world-knowledge question answering (QA), and episodic memory, but a correct answer does not show that the agent ever observed the evidence for it, and progress does not show what the agent learned. We introduce \bench{}, which measures one agent's self-improvement through three views on the same run: what it knows, whether its own evidence supports that knowledge, and how far it gets in the world. It is built on the game Outer Wilds, where each episode is a 22-minute loop after which the world resets, only the agent's knowledge carries over, and progress is gated by knowledge. After every episode, we measure the places reached and ship-log rumor facts revealed, the correctness of answers to 164 questions about the world, and whether each answer is supported by the frames and position the agent cites from its own recorded play. Two LLM judges from different model families check every answer. In eight-episode baselines with \gpt{} and \gem{} planners, which act through a learned controller or their own gamepad inputs, the three views do not move together, so any one of them alone gives an incomplete picture of whether the agent is improving. The \gpt{} agent's correct answers grow from 14.6\% to 26.8\% of all questions, but at episode 8 only 59 to 68\% of them are supported by its cited evidence. It reveals no new rumor facts after episode 3, while the places it reaches and its memory keep growing. More than half of the answers that become correct after episode 3 cite only evidence from earlier episodes.
\end{abstract}

\section{Introduction}

LLM agents are increasingly expected to improve themselves from their own experience across episodes \citep{shinn2023reflexion,wang2023voyager,agentodyssey2026}. By self-improvement we mean this kind of improvement, with fixed weights, close to what AgentOdyssey calls test-time continual learning, not a model training on its own outputs. An agent can also improve how it acts, by making its controller execute what it intends more reliably. In this version the controller is fixed, and we leave its online improvement to future work. Benchmarks already evaluate parts of it. Game and embodied benchmarks measure task progress and exploration \citep{zhang2025videogamebench,ouyang2026gameworld,ju2026mineexplorer}. AgentOdyssey measures game progress together with world-knowledge question answering (QA), episodic-memory QA, and exploration coverage in generated text games \citep{agentodyssey2026}, and ExplorationBench measures how agents acquire knowledge by experimenting in symbolic worlds \citep{zhang2026explorationbench}. Embodied QA asks agents questions about environments they explore \citep{das2018eqa,majumdar2024openeqa}.

In an embodied world, an agent that improves itself should show it in three ways. It should know more about the world. It should be able to back what it knows with its own experience, pointing to when and where it learned each fact. And it should reach places and information it could not reach before. These are views of the same improvement, but they can come apart, so one of them alone can misread it. A correct answer does not show that the agent observed the evidence for it. In embodied QA, an agent that never reaches a target can still answer 45.5\% of questions, about as many as without any images \citep{garg2026answering}. Memory can grow without new discoveries. And progress depends on actions that may not execute as intended. Grounded QA and attribution work show that answers and their supporting evidence need to be checked separately \citep{rashkin2023ais,xiao2024nextgqa}. For an agent that explores on its own, the evidence also has to be found in the experience the agent itself collected.

\begin{figure*}[t]
\centering
\includegraphics[width=\linewidth]{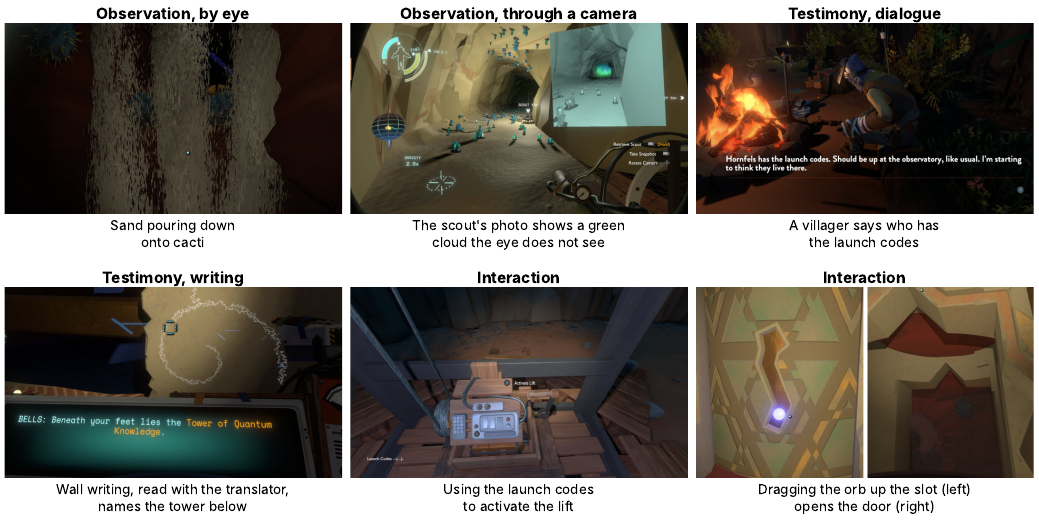}
\caption{Types of evidence an agent can ground knowledge in: observation (by eye or through an instrument), testimony (dialogue or writing, shown as on-screen text), and interaction, the result of the agent's own action \citep{gopnik2004theory,harris2006trust}. Screenshots from Outer Wilds, \textcopyright{} Mobius Digital.}
\label{fig:sources}
\end{figure*}

Outer Wilds \citep{outerwilds2019} lets us take the three views on the same run. Its episodes are loops of about 22 minutes. Each loop ends when the Sun goes supernova, the solar system resets, and only the player's knowledge carries over, as in a multi-episode trial in meta-reinforcement learning \citep{duan2016rl2}. What makes it useful is what happens inside each loop. The game defines no explicit task reward. Its progress signals are sparse and implicit, revealing ship-log rumors and reaching places, and most of them are gated by knowledge acquired elsewhere. The knowledge is of different types (Table~\ref{tab:categories}). Temporal knowledge decides when a place opens, such as the Ash Twin Project, which can be entered only late in the loop. Spatial and temporal knowledge decide whether the agent can return, since rising sand closes the way out of the Sunless City. Knowledge of function opens routes that look closed, since jellyfish carry the agent through an electrical layer. Knowledge of appearance keeps the agent alive, since ghost matter is invisible to the eye but shows in scout photos. Narrative knowledge explains who built these places, what happened to them, and why, and procedural knowledge, how to reach a place, combines the others. This knowledge comes from evidence of three types (Figure~\ref{fig:sources}): observation, what the agent sees, testimony, what it reads or is told, and interaction, what it learns from the results of its own actions \citep{gopnik2004theory,harris2006trust}. Collecting that evidence is part of the task. The world changes on its own schedule within each episode, so evidence reachable at one time may be closed later, and actions run through imperfect control.

We introduce \bench{} to measure self-improvement through these three views. After every episode, we measure progress as the cumulative number of places reached and rumor facts revealed, recorded by a game-side mod and never shown to the agent. We measure knowledge with \qa{}, 164 single-fact questions written from the game wiki. And we measure grounding by requiring every answer to cite the agent's own experience: the loop and frame range where it learned the answer, and the body and $x,y,z$ where it was. In the terms of \citet{tulving1972episodic}, the answer is semantic knowledge and the citation is episodic memory. An answer counts only when it is correct and attributable to the cited evidence \citep{rashkin2023ais}, as judged by two LLM judges from different model families.

\begin{figure}[t]
\centering
\includegraphics[width=\columnwidth]{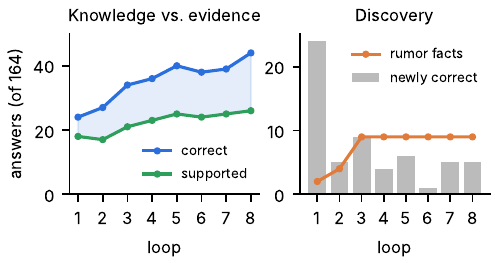}
\caption{\gpt{} with the knowledge base, \gpt{} judge. Left: correct and grounded answers. Right: newly correct answers per loop (bars) and cumulative rumor facts (line).}
\label{fig:divergence}
\end{figure}

We run baselines with \gpt{} and \gem{} planners over eight episodes, acting through a learned controller or their own gamepad inputs. The three views do not move together (Figure~\ref{fig:divergence}). The \gpt{} agent's correct answers rise overall across the eight loops, but its grounded answers grow more slowly. It reveals no rumor facts after loop 3, while the places it reaches grow from 44 to 57 and its memory from 618 to 1{,}046 records. And 12 of the 21 answers that become correct after loop 3 cite only evidence from earlier loops.

\paragraph{Contributions.}
\begin{enumerate}[leftmargin=*,itemsep=2pt]
  \item \bench{}, a benchmark that measures an agent's self-improvement across episodes through knowledge, grounding, and progress on the same run, with the number of episodes as the primary cost, in a world where only knowledge survives each episode and progress is gated by knowledge.
  \item \qa{}, an evidence-grounded QA protocol that ties every answer to the agent's own episodic record, judged for correctness and attribution by two LLM judges whose agreement we report.
  \item Baselines over eight episodes and an analysis of the evidence each answer cites, which show that rumor discovery stops early while correct answers rise overall and places and memory keep growing, and that grounded answers grow more slowly than correct ones.
\end{enumerate}

\section{Related Work}

\paragraph{Exploration and learning across episodes.}
Meta-reinforcement learning keeps memory across the episodes of a trial and reports performance as the trial goes on \citep{duan2016rl2,bauer2023ada}. For LLM agents, AgentOdyssey evaluates test-time continual learning in generated text games with game progress, world-knowledge QA, episodic-memory QA, and exploration coverage \citep{agentodyssey2026}. ExplorationBench measures how agents acquire knowledge by experimenting in symbolic worlds whose rules contradict familiar knowledge \citep{zhang2026explorationbench}, and AdaExploreBench tests exploration when hidden rules drift \citep{wang2026adaexplorebench}. DiscoveryWorld scores task completion, task-relevant actions, and the explanatory knowledge an agent discovers in a text-based world \citep{jansen2024discoveryworld}, and UltraDiscovery asks agents to revise a taught theory and predict an unseen intervention across interconnected domains \citep{li2026ultradiscovery}. MineExplorer evaluates visual agents exploring Minecraft with milestone checks \citep{ju2026mineexplorer}, and ASIWorld evaluates agents in a persistent survival sandbox by task success, development, and survival costs \citep{lin2026asiworld}. As in meta-reinforcement learning, we take the number of episodes as the cost. Next to progress and knowledge, we check whether each answer is supported by evidence the agent cites from its own play, in a 3D world that resets every loop and is acted in through imperfect control.

\paragraph{Embodied QA and evidence.}
Embodied QA asks agents to answer questions about an environment they explore, and OpenEQA covers both episodic-memory and active settings \citep{das2018eqa,majumdar2024openeqa}. Long-term active EQA combines memory of past observations with new exploration to answer time-dependent questions \citep{ginting2025mindpalace}. EXPRESS-Bench scores whether an answer is grounded in what the agent observed together with the reliability of its exploration \citep{jiang2025express}. \citet{garg2026answering} vary how much evidence the answerer gets and show that correct embodied answers often do not come from collected evidence, PQB-EQA pairs each question with two environments that need different answers so that guessing cannot pass \citep{shelton2025pqb}, and AbstainEQA tests whether agents abstain when their observations are insufficient \citep{wu2026abstaineqa}. Attribution work asks whether text is supported by a cited source \citep{rashkin2023ais,gao2023alce}, and NExT-GQA asks video QA models to ground answers in a time interval \citep{xiao2024nextgqa}. In question-directed embodied QA, the agent explores to answer given questions. Our agent explores on its own, is asked the same questions after every episode, and cites its own evidence, and we track how often its answers are supported over successive loops. LLM judges are widely used \citep{zheng2023judging} but can favor their own model family \citep{panickssery2024llm}, so we use two families \citep{verga2024replacing}.

\paragraph{Memory for agents.}
LLM agents improve over trials with verbal reflection \citep{shinn2023reflexion}, memory streams \citep{park2023generative}, or memory reused across task streams \citep{wei2025evomemory}. AriGraph builds a knowledge graph with episodic memory while an agent explores text games \citep{anokhin2024arigraph}. Zep keeps a temporal knowledge graph whose facts link to their source episodes and record when each fact holds \citep{rasmussen2025zep}, EnSIMem indexes conversation episodes by entity records \citep{meng2026ensimem}, and BeliefMem keeps competing beliefs with probabilities \citep{liao2026beliefmem}. Our baseline uses an entity-centric knowledge base with evidence-linked, append-only records, as a practical baseline for the benchmark, not a proposed memory method.

\paragraph{Planning and execution.}
SayCan grounds an LLM planner in what its robot skills can achieve \citep{ahn2022saycan}, LITEN lets a planner learn at inference time what its controller can do \citep{shah2025liten}, and STAGE shows that embodied agents can understand an instruction yet fail to act on it \citep{jin2026stage}. VPT and NitroGen learn game control from human play \citep{baker2022vpt,magne2026nitrogen}, and several works relabel demonstrations with language \citep{andrychowicz2017her,lynch2021language,xiao2023dial,zhang2023sprint}. In our setting, an execution failure can look like a knowledge failure, because it changes what evidence the agent collects. This version does not separate the two.

\section{Environment}
\label{sec:env}

Outer Wilds is a small solar system of hand-built planets. Each loop ends when the Sun goes supernova, about 22 minutes of game time after waking, or when the player dies, and the next loop starts at the same campfire with every physical change undone. Within a loop, the planets follow their own schedule whether the player is there or not. Sand flows from Ash Twin to Ember Twin, and the crust of Brittle Hollow breaks apart. Some consequences cannot be undone within the loop, and the cost of an action depends on when it occurs. The game has no levels, items, or experience points. Its ship log records rumors, short facts about places the player has heard of. Table~\ref{tab:scenarios} lists scenarios documented in the game wiki \citep{outerwildswiki} that make the world hard for an agent.

We built a game-side mod that records every frame with the gamepad inputs, whether the player is on foot or in the ship, the body the player is on or nearest to, the position $x,y,z$ relative to that body, the velocity, facing direction, and angular velocity, and the game time of the step. It can freeze and resume the game, and it records when the player enters one of the game's named areas and when a ship-log fact is revealed. These events go to the scorer only and are never shown to the agent.

\begin{table*}[t]
\centering
\small
\begin{tabular}{p{0.5\textwidth}p{0.18\textwidth}p{0.24\textwidth}}
\toprule
Scenario & Knowledge needed & Property \\
\midrule
The Ash Twin Project can be entered only from one tower on Ash Twin, after the sand has drained enough to uncover it, and only while Ember Twin is directly overhead. The earliest entry is about 8 minutes after waking. & Temporal, procedural & Open only in a time window \\
Sand flows from Ash Twin to Ember Twin over the loop. The lowest district of the Sunless City on Ember Twin is blocked early, and a player inside can be trapped by the rising sand. & Spatial, temporal & Irreversible within the loop \\
An electrical layer below the ocean current of Giant's Deep blocks its core. Jellyfish insulate against electricity, so a player inside one is carried through. & Function, procedural & Knowledge-gated access \\
Most tornadoes on Giant's Deep throw the ship up and away. A tornado that spins the other way pulls it below the ocean current. & Causal & Similar look, opposite effect \\
Brittle Hollow's crust breaks apart and falls into its black hole during the loop, and routes through the Hanging City deteriorate. A player who falls in comes out of a white hole near the White Hole Station, which mainly costs time. & Spatial, causal & The world changes on its own schedule \\
Ghost matter cannot be seen by eye but shows in scout photos, and walking into it damages or kills the player. & Appearance, causal & Visible only through an instrument \\
The sand pillar between the twins can catch the player or the ship and carry it toward Ember Twin, off its intended course. & Causal & The environment can override control \\
\bottomrule
\end{tabular}
\caption{Challenging scenarios in the world of \bench{}, with the knowledge types of Table~\ref{tab:categories}. Each property makes evidence harder to collect or progress harder to make: a time window or an irreversible choice limits when the agent can act, a knowledge-gated route stays closed until the agent learns how to pass it, the world changing on its own makes one visit insufficient, similar-looking things and hidden hazards make a single observation misleading, and the environment can make an action differ from the intended one.}
\label{tab:scenarios}
\end{table*}

\section{Evaluation}

\subsection{Progress}

The primary cost is the number of loops, and the secondary cost is game time. The game clock runs only while the agent acts, not while it plans. We measure progress by two counts. A place is one of the game's named areas, counted the first time the player enters it in the run, except the areas the player wakes up in. A rumor fact is a ship-log rumor, counted the first time it is revealed in the run, so facts already revealed in the starting save do not count. For both, we report the cumulative count after each loop and the sum of these counts over all loops. The sum is the area under the cumulative discovery curve. It rewards an agent that reveals early over one that reveals the same amount only at the end, and it needs no optimal reference policy.

\subsection{\qa}

\paragraph{Questions.}
We wrote 164 single-fact questions about the base game from the Outer Wilds Wiki \citep{outerwildswiki}, without consulting any agent's records. Questions use only names the game presents to the player. We label each question with one of eight knowledge types by what it asks about the world (Table~\ref{tab:categories}). The types are built on the question categories of OpenEQA \citep{majumdar2024openeqa} and NExT-QA \citep{xiao2021nextqa}, the commonsense dimensions of \citet{ilievski2021dimensions}, and procedural knowledge in the sense of \citet{anderson1983architecture}, as Table~\ref{tab:categories} shows. Narrative covers the world's story: past events, who made things, and why. OpenEQA has no such category, since its household scenes have no story. The labels have not been checked by a second annotator. Evidence types belong to answers, not questions (Figure~\ref{fig:sources}). We distinguish observation, testimony, and interaction, the sources from which children learn about the world in developmental research: what they observe and what they find by acting on the world, which \citet{gopnik2004theory} call intervention, and what others tell them \citep{harris2006trust}. The questions do not yet cover metacognitive knowledge, such as the agent's uncertainty about its own beliefs.

\begin{table*}[t]
\centering
\small
\begin{tabular}{lp{0.345\textwidth}p{0.175\textwidth}rcc}
\toprule
Knowledge type & Example (reference) & Corresponds to & $n$ & Correct & Grounded \\
\midrule
Appearance & What is Slate wearing on their head? (A welding mask.) & Attribute recognition, quality & 18 & 5 & 3 \\
Spatial & Where is Hornfels usually found? (At the observatory.) & Object localization, spatial & 20 & 4 & 4 \\
Identity & Who has the launch codes? (Hornfels.) & Object recognition, taxonomic & 36 & 14 & 10 \\
Function & What do the launch codes let the player do? (Activate the lift.) & Utility & 19 & 7 & 4 \\
Causal & What happens if you walk into ghost matter? (It damages you.) & Causal & 14 & 5 & 3 \\
Temporal & About how long after you wake up does the Sun go supernova? (About 22 minutes.) & Temporal & 6 & 3 & 0 \\
Narrative & Why did the Nomai come to this solar system? (They followed a signal.) & Creation, desire & 41 & 6 & 2 \\
Procedural & How can you get past the electrical layer? (Ride inside a jellyfish.) & Functional reasoning, procedural & 10 & 0 & 0 \\
\bottomrule
\end{tabular}
\caption{Questions by knowledge type, with the corresponding categories of OpenEQA \citep{majumdar2024openeqa} and NExT-QA \citep{xiao2021nextqa} and dimensions of \citet{ilievski2021dimensions}, and the \gpt{} run's correct and grounded answers at loop 8 (\gpt{} judge).}
\label{tab:categories}
\end{table*}

\paragraph{Answering.}
To evaluate loop $L$, the tested LLM answers from its memory at the end of loop $L$, and may request memory records and frames from loops $1$ to $L$, never later. It cites up to three pieces of evidence, each with loop and frame range, body and $x,y,z$, and a short source, or replies ``I don't have enough evidence.'' A single piece of evidence may not establish every part of an answer. An appearance question may need evidence of both which entity it is and how it looks, and a narrative question may depend on dialogue or writing.

\paragraph{Correctness and grounding.}
A program first checks each cited piece: the frames exist in a loop no later than $L$, the cited body matches the recording, and the cited position is within 30 units of the recorded position at the first cited frame. The correctness judge compares the answer with the reference, and labels it correct, wrong, or no answer. The agent may use its own name or description for a thing if it clearly identifies that one thing. We call an answer grounded when it is attributable to the agent's own episodic record \citep{rashkin2023ais}, where the sources are the cited frames and the recorded position. For correct answers, the grounding judge sees the cited frames, the recorded body, position, and game time, and the program checks, but not the reference. It decides whether the evidence shows the answer and ties it to the thing the question names. The judge may request more frames, but frames outside the cited ranges cannot supply a missing fact, except the ``Talk to'' prompt at the start of a cited conversation.

\paragraph{Score.}
A question scores $+1$ if correct and grounded, $-1$ if correct but not grounded, and $0$ if wrong or unanswered. The score is the sum over 164 questions scaled to 100. A correct answer that the agent cannot ground counts against it, because its account of how it learned the fact does not hold up, so the score penalizes unsupported claims.

\paragraph{Judges.}
One \gpt{} and one \gem{} judge, the same models as the planners, each judge every answer, and we report both. We measure their agreement with Cohen's $\kappa$ \citep{cohen1960kappa}, counting unanswered questions as the same decision.

\section{Baseline Agent}

The baseline agent has three parts: an LLM planner that decides, an entity-centric knowledge base that carries what it learned across loops, and a hybrid action interface, in which an action is carried out either by a learned controller or by gamepad inputs the planner writes itself. The two conditions use the same planner and controller, and both keep the action history and earlier frames. They differ in the knowledge base, and the run with the knowledge base is also told that it will be evaluated with \qa{} (Table~\ref{tab:conditions}).

\begin{table}[t]
\centering
\small
\begin{tabular}{lcc}
\toprule
Knowledge base & With & Without \\
\midrule
Last action's frames and log & \checkmark & \checkmark \\
Actions in this loop, with verdicts & \checkmark & \checkmark \\
Request frames of earlier loops & \checkmark & \checkmark \\
Entity list, record lookup & \checkmark & \\
Writes records & \checkmark & \\
Scored by \qa{} & \checkmark & \\
\bottomrule
\end{tabular}
\caption{What the planner has in each condition, besides the current view and its previous plan. Each planner is told how its run is evaluated. Both runs are evaluated by discoveries and places, and only the run with the knowledge base also with \qa{}.}
\label{tab:conditions}
\end{table}

\paragraph{Planner.}
The planner is \gpt{} \citep{openai2026gpt61sol} or \gem{} \citep{google2026gemini4argon}, through the APIs gpt-6.1-sol and gemini-4-argon, without fine-tuning, and the game is frozen while it thinks. Neither the planner nor the controller receives audio. At each step it gets the current view, its last action with what it expected and how the action ended, the frames and action log of that action, its recent actions in this loop with its own verdicts, and its previous plan. The frames of an action are its first and last frames and up to four frames where the view changed most. The action log gives the inputs sent and the recorded state at every frame (Section~\ref{sec:env}). Before deciding, it may request any frame of this or an earlier loop. It then judges whether its last action succeeded, keeps or revises its plan, a list of goals broken into steps, describes the current view, and writes the next action as one first-person instruction. Like the controller's labels, its descriptions and actions name things by how they look, using game names only for planets, tools, and text written in the view. The planner is told the goal of the run (explore, learn about the world, keep the player safe) and how it will be evaluated. It is not told that reading or talking gives information, how long a loop lasts, or how many loops it has.

\paragraph{Two ways to act.}
An action is carried out in one of two ways. The controller does actions that need looking while acting, such as walking to something, flying, or landing. For exact or discrete inputs, such as a button shown in an on-screen prompt, a dialogue option, or a short exact turn, the planner writes the gamepad inputs itself, frame by frame, for at most 170 frames (about 9.5 seconds of game time), and sees the view again only when they end. In the \gpt{} run with the knowledge base, the controller carried out 501 of 2,509 actions (20\%), and 760 of 3,347 (23\%) without the knowledge base. In the \gem{} run it carried out 27 of 930 actions in loops 1 to 7 (3\%).

\paragraph{Entity-centric knowledge base.}
Each record in the knowledge base is an entity, its type, property--value pairs, the frames that support it (for example, loop 2, frames 300--340), and a short note on how the planner learned it. The program adds the loop and action in which the record was written and the recorded body and $x,y,z$ position of its first cited frame. After each action, in the same decision in which it keeps or revises its plan, the planner looks at the action's frames and log and adds a record for every explicit thing it learned. A fact seen again is added as a new record of the same entity with its own frames, so evidence for a claim accumulates instead of being merged. At every step it sees the list of its entity names with their record counts, and it can request the records of any entity. In records only, it calls an entity by its real name once it has learned the name, and by its appearance until then. Records are only added, never changed or deleted. When records contradict each other, the planner is asked to compare their frames and add a contradiction record that marks records as accepted, rejected, or undecided, with a rationale, which may also explain how both claims can hold, for example at different times in the loop. Each loop starts from the previous loop's final knowledge base. The \gpt{} run's knowledge base grows to 1{,}046 records about 40 entities by loop 8, and the planner never wrote a contradiction record. Questions about loop $L$ use only the records written by loop $L$.

\paragraph{Controller.}
The controller is trained by imitation on 5 hours of human gameplay, cut into 8,436 atomic actions. Each action is one intent, such as walking up to something or landing the ship, with a goal-level label that names things by how they look and is understandable from the action's first frame alone. Reference labels were written by human annotators, edited by an LLM for grammar and clarity, and decided by a human. Because a planner may describe the same scene in other words than the annotators, each tested LLM relabels every action: it first sees only the action's first frame, as when playing, and proposes candidate actions, then sees the reference label and the whole action and writes its final label in its own words. Each LLM's controller is trained on its own version, conditioned on its description and label, and outputs the gamepad's analog inputs and 13 buttons until an end-of-action token. It is trained without reinforcement learning, to execute the given intent faithfully, so that a bad decision stays visible as the planner's. Its design follows VPT \citep{baker2022vpt}. A frozen ImageNet ResNet-34 encodes each frame at full resolution, and the features are conditioned on the text with FiLM \citep{perez2018film} and averaged into 16 region tokens per frame. A frozen T5-base encoder \citep{raffel2020t5} encodes the label and the description. A six-layer transformer attends over the recent steps and the text and predicts binned stick and trigger positions, 13 buttons, and the end of the action. An action ends when the end-of-action probability is above 0.4 on six steps in a row, or after 170 steps of 0.056 seconds, with the game frozen between steps. It has 26.8M trainable parameters besides the frozen ResNet-34 and T5-base encoder. Each planner's controller was trained on its own GPU with 180 GB of memory, for two epochs (17,148 steps with a window of 32 steps, about two hours). The controller is a simple baseline and is not yet reliable.

\section{Results}

We ran three conditions for eight loops each, one run per condition: \gpt{} with the knowledge base, \gpt{} without the knowledge base, and \gem{} with the knowledge base. \qa{} applies to the two runs with the knowledge base.

\begin{table}[t]
\centering
\small
\resizebox{\columnwidth}{!}{%
\begin{tabular}{lcccc}
\toprule
& \multicolumn{2}{c}{Places} & \multicolumn{2}{c}{Rumor facts} \\
\cmidrule(lr){2-3}\cmidrule(lr){4-5}
Run & At loop 8 & Sum over loops & At loop 8 & Sum over loops \\
\midrule
\gpt{}, with knowledge base & 57 & 346 & 9 & 60 \\
\gpt{}, without knowledge base & 47 & 272 & 6 & 37 \\
\gem{}, with knowledge base & 30 & 172 & 1 & 6 \\
\bottomrule
\end{tabular}}
\caption{Progress. The sum adds the cumulative count after each of the eight loops.}
\label{tab:exploration}
\end{table}

\begin{figure}[t]
\centering
\includegraphics[width=\columnwidth]{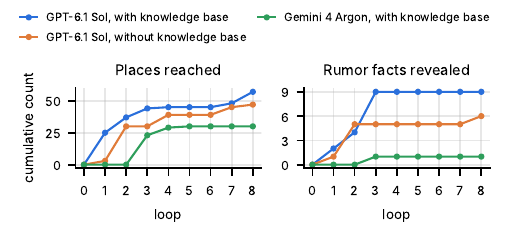}
\caption{Places reached and rumor facts revealed, cumulative over eight loops.}
\label{fig:exploration}
\end{figure}

\paragraph{Progress.}
\gpt{} with the knowledge base reaches the most places and rumor facts, at loop 8 and summed over loops (Table~\ref{tab:exploration}, Figure~\ref{fig:exploration}). \gpt{} with the knowledge base reveals 9 rumor facts by loop 3 and none after. Its places keep growing, from 44 at loop 3 to 57 at loop 8, and its knowledge base grows from 618 records at loop 3 to 1{,}046 at loop 8.

\begin{table*}[t]
\centering
\small
\setlength{\tabcolsep}{4pt}
\begin{tabular}{c r cc cc rr}
\toprule
& & \multicolumn{2}{c}{Correct (\%)} & \multicolumn{2}{c}{Grounded (\%)} & \multicolumn{2}{c}{Score} \\
\cmidrule(lr){3-4}\cmidrule(lr){5-6}\cmidrule(lr){7-8}
Loop & Answered & \gpt{} & \gem{} & \gpt{} & \gem{} & \gpt{} & \gem{} \\
\midrule
\multicolumn{8}{l}{\textit{\gpt{} planner, with knowledge base}} \\
1 & 27 & 14.6 & 14.6 & 11.0 & 11.6 & 7.3 & 8.5 \\
2 & 29 & 16.5 & 15.2 & 10.4 & 9.8 & 4.3 & 4.3 \\
3 & 39 & 20.7 & 21.3 & 12.8 & 15.2 & 4.9 & 9.1 \\
4 & 39 & 22.0 & 22.0 & 14.0 & 15.9 & 6.1 & 9.8 \\
5 & 43 & 24.4 & 23.8 & 15.2 & 15.9 & 6.1 & 7.9 \\
6 & 40 & 23.2 & 22.6 & 14.6 & 15.9 & 6.1 & 9.1 \\
7 & 42 & 23.8 & 24.4 & 15.2 & 17.1 & 6.7 & 9.8 \\
8 & 48 & 26.8 & 26.8 & 15.9 & 18.3 & 4.9 & 9.8 \\
\midrule
\multicolumn{8}{l}{\textit{\gem{} planner, with knowledge base}} \\
1 & 2 & 0.6 & 0.6 & 0.6 & 0.6 & 0.6 & 0.6 \\
2 & 1 & 0.6 & 0.6 & 0.6 & 0.6 & 0.6 & 0.6 \\
3 & 6 & 3.0 & 3.7 & 1.8 & 3.0 & 0.6 & 2.4 \\
4 & 12 & 7.3 & 7.3 & 3.7 & 6.7 & 0.0 & 6.1 \\
5 & 16 & 8.5 & 8.5 & 3.7 & 6.7 & $-$1.2 & 4.9 \\
6 & 13 & 7.9 & 7.9 & 3.0 & 6.7 & $-$1.8 & 5.5 \\
7 & 17 & 10.4 & 10.4 & 3.7 & 7.9 & $-$3.0 & 5.5 \\
8 & 14 & 7.9 & 8.5 & 3.7 & 6.7 & $-$0.6 & 4.9 \\
\bottomrule
\end{tabular}
\caption{\qa{} per loop under the \gpt{} and \gem{} judges. Correct and grounded are shares of all 164 questions. The score runs from $-100$ to $100$.}
\label{tab:qa}
\end{table*}

\paragraph{Knowledge and grounding.}
Table~\ref{tab:qa} gives the QA results. The \gpt{} run's correct answers grow from 14.6\% of all questions at loop 1 to 26.8\% at loop 8, and the \gem{} run's from 0.6\% to about 8\%. When the agents answer, they are mostly right: at loop 8, 44 of the \gpt{} run's 48 answers are correct. Grounded answers grow more slowly. At loop 8, only 26 to 30 of the 44 correct answers are grounded in the agent's own evidence, depending on the judge. By knowledge type (Table~\ref{tab:categories}), the \gpt{} agent grounds all 4 of its correct spatial answers and 10 of its 14 correct identity answers, but only 2 of its 6 correct narrative answers, and none of its 3 correct temporal answers. No run answers a procedural question correctly. These counts are small and from one run.

\begin{table}[t]
\centering
\small
\resizebox{\columnwidth}{!}{%
\begin{tabular}{lcccccccc}
\toprule
Loop & 1 & 2 & 3 & 4 & 5 & 6 & 7 & 8 \\
\midrule
Correct answers & 24 & 27 & 34 & 36 & 40 & 38 & 39 & 44 \\
Newly correct & 24 & 5 & 9 & 4 & 6 & 1 & 5 & 5 \\
\quad citing only earlier loops & -- & 1 & 1 & 3 & 4 & 1 & 2 & 2 \\
\bottomrule
\end{tabular}}
\caption{Which loops the \gpt{} run's correct answers cite (\gpt{} judge). An answer cites only earlier loops when its newest evidence is from a loop before the evaluated one.}
\label{tab:provenance}
\end{table}

\paragraph{Where knowledge comes from.}
Table~\ref{tab:provenance} shows which loops the correct answers cite. At loop 8, 34 of the \gpt{} run's 44 correct answers cite evidence from loops 1 to 3, and 21 cite loop 1 alone. From loop 4 on, 12 of the 21 newly correct answers cite only earlier loops. So more than half of the gains in later loops come from answers backed by evidence the agent already had, not from evidence collected in that loop. The citation shows where the supporting evidence is, not when the agent first learned the fact.

\paragraph{Judge agreement.}
The judges agree on 97.5\% of questions in the \gpt{} run ($\kappa = 0.93$) and 97.6\% in the \gem{} run ($\kappa = 0.80$), and almost never disagree on correctness ($\kappa = 0.97$ and $0.99$). These numbers are high partly because most questions are unanswered. On whether a correct answer is grounded, they agree much less in the \gem{} run (61.8\%, $\kappa = 0.29$, on 76 answers) than in the \gpt{} run (92.4\%, $\kappa = 0.83$, on 275 answers), and the \gpt{} judge is always the stricter one.

\section{Conclusion}

In our baselines, the three views of self-improvement that \bench{} measures do not move together. By correctness alone, the agent looks more improved than its own evidence supports, and by rumor discovery alone, less.

\section*{Limitations}

\paragraph{Runs.} Each condition is a single run, so we report observations, not effects. The two \gpt{} runs differ both in the knowledge base and in whether the planner is told about \qa{}, so their difference in progress cannot be attributed to the knowledge base alone. We do not yet show why rumor discovery stops. It could come from planning, from the controller, or from the time budget, and separating them needs conditions with given knowledge or given control.

\paragraph{Controller data and the intent--action gap.} The controller is trained on 8,436 actions from about 5 hours of play. This is not enough, and the actions are not evenly distributed across the world and its kinds of action. The controller is not yet reliable, and the planner carried out most actions with its own gamepad inputs. Per-planner relabeling has not been compared with training on the reference labels alone. Future versions will scale the controller data and balance it, and compare different controllers with the planner held fixed. The controller is also fixed during play. Self-improvement can include the controller itself, learning online from its own attempts to execute what the planner intends, so that the gap between intended and executed actions shrinks. We expect future methods to improve both progress and the controller online, with the number of online actions as a cost. Measuring this will need a check of whether each action does what its instruction says, kept separate from actions the planner carries out itself.

\paragraph{Judges.} The judges are LLMs, and their agreement with each other does not show that they are right. They were not compared with human judges, so we do not know how often they wrongly accept or reject an answer. They do not receive reference information about what each entity looks like, so a correct answer can fail grounding when the cited frames show the entity but do not name it. By their stated reasons, the judges disagree in the \gem{} run because evidence cited while the agent was moving fails the position check at the first cited frame, because they handle one failed piece of evidence differently, and because only one judge asks for more frames when a cited range is long. A grounded answer cites evidence that supports it, but this does not show that the agent gave the answer because of that evidence \citep{shelton2025pqb}. Future versions will compare the judges with human judges on a sample of answers, give them reference information about each entity, and check positions over the whole cited range instead of only its first frame.

\paragraph{Questions.} The questions are not balanced across knowledge types. Narrative and identity questions make up 77 of the 164, while temporal has 6 and procedural 10. Future versions will write questions for each type, with enough of each to compare them, and label the evidence each answer cites by type. Scoring is binary, so a true but less specific answer can be judged wrong. Both planners may know Outer Wilds from pretraining. The grounding check reduces this problem but does not remove it.

\bibliography{references}

\begin{thebibliography}{54}
\providecommand{\natexlab}[1]{#1}

\bibitem[{{Adaptive Agent Team} et~al.(2023){Adaptive Agent Team}, Bauer,
  Baumli et~al.}]{bauer2023ada}
{Adaptive Agent Team}, Jakob Bauer, Kate Baumli, et~al. 2023.
\newblock Human-timescale adaptation in an open-ended task space.
\newblock In \emph{Proceedings of the 40th International Conference on Machine
  Learning}.

\bibitem[{Ahn et~al.(2022)Ahn, Brohan, Brown et~al.}]{ahn2022saycan}
Michael Ahn, Anthony Brohan, Noah Brown, et~al. 2022.
\newblock Do as {I} can, not as {I} say: Grounding language in robotic
  affordances.
\newblock In \emph{Conference on Robot Learning}.

\bibitem[{Anderson(1983)}]{anderson1983architecture}
John~R. Anderson. 1983.
\newblock \emph{The Architecture of Cognition}.
\newblock Harvard University Press.

\bibitem[{Andrychowicz et~al.(2017)Andrychowicz, Wolski, Ray, Schneider, Fong,
  Welinder, McGrew, Tobin, Abbeel, and Zaremba}]{andrychowicz2017her}
Marcin Andrychowicz, Filip Wolski, Alex Ray, Jonas Schneider, Rachel Fong,
  Peter Welinder, Bob McGrew, Josh Tobin, Pieter Abbeel, and Wojciech Zaremba.
  2017.
\newblock Hindsight experience replay.
\newblock In \emph{Advances in Neural Information Processing Systems}.

\bibitem[{Anokhin et~al.(2024)Anokhin, Semenov, Sorokin, Evseev, Kravchenko,
  Burtsev, and Burnaev}]{anokhin2024arigraph}
Petr Anokhin, Nikita Semenov, Artyom Sorokin, Dmitry Evseev, Andrey Kravchenko,
  Mikhail Burtsev, and Evgeny Burnaev. 2024.
\newblock {AriGraph}: Learning knowledge graph world models with episodic
  memory for {LLM} agents.
\newblock \emph{arXiv preprint arXiv:2407.04363}.

\bibitem[{Baker et~al.(2022)Baker, Akkaya, Zhokhov, Huizinga, Tang, Ecoffet,
  Houghton, Sampedro, and Clune}]{baker2022vpt}
Bowen Baker, Ilge Akkaya, Peter Zhokhov, Joost Huizinga, Jie Tang, Adrien
  Ecoffet, Brandon Houghton, Raul Sampedro, and Jeff Clune. 2022.
\newblock Video pretraining ({VPT}): Learning to act by watching unlabeled
  online videos.
\newblock In \emph{Advances in Neural Information Processing Systems}.

\bibitem[{Cohen(1960)}]{cohen1960kappa}
Jacob Cohen. 1960.
\newblock A coefficient of agreement for nominal scales.
\newblock \emph{Educational and Psychological Measurement}, 20(1):37--46.

\bibitem[{Das et~al.(2018)Das, Datta, Gkioxari, Lee, Parikh, and
  Batra}]{das2018eqa}
Abhishek Das, Samyak Datta, Georgia Gkioxari, Stefan Lee, Devi Parikh, and
  Dhruv Batra. 2018.
\newblock Embodied question answering.
\newblock In \emph{Proceedings of the IEEE Conference on Computer Vision and
  Pattern Recognition}.

\bibitem[{Duan et~al.(2016)Duan, Schulman, Chen, Bartlett, Sutskever, and
  Abbeel}]{duan2016rl2}
Yan Duan, John Schulman, Xi~Chen, Peter~L. Bartlett, Ilya Sutskever, and Pieter
  Abbeel. 2016.
\newblock {RL}$^2$: Fast reinforcement learning via slow reinforcement
  learning.
\newblock \emph{arXiv preprint arXiv:1611.02779}.

\bibitem[{Gao et~al.(2023)Gao, Yen, Yu, and Chen}]{gao2023alce}
Tianyu Gao, Howard Yen, Jiatong Yu, and Danqi Chen. 2023.
\newblock Enabling large language models to generate text with citations.
\newblock In \emph{Proceedings of the 2023 Conference on Empirical Methods in
  Natural Language Processing}.

\bibitem[{Garg et~al.(2026)Garg, Pandey, Rawal, Khandelwal, and
  Sheth}]{garg2026answering}
Shubh Garg, Divyansh Pandey, Niyati Rawal, Vedant Khandelwal, and Amit Sheth.
  2026.
\newblock Answering without arriving: An evidence audit for embodied question
  answering.
\newblock In \emph{NeurIPS 2026 Workshop on Neuro-Symbolic Embodied
  Intelligence}.

\bibitem[{Ginting et~al.(2025)Ginting, Kim, Meng, Reinke, Krishna, Kayhani,
  Peltzer, Fan, Shaban, Kim, Kochenderfer, Agha-mohammadi, and
  Omidshafiei}]{ginting2025mindpalace}
Muhammad~Fadhil Ginting, Dong-Ki Kim, Xiangyun Meng, Andrzej~Marek Reinke,
  Bandi~Jai Krishna, Navid Kayhani, Oriana Peltzer, David Fan, Amirreza Shaban,
  Sung-Kyun Kim, Mykel Kochenderfer, Ali-akbar Agha-mohammadi, and Shayegan
  Omidshafiei. 2025.
\newblock Enter the mind palace: Reasoning and planning for long-term active
  embodied question answering.
\newblock In \emph{Proceedings of The 9th Conference on Robot Learning}, volume
  305 of \emph{PMLR}, pages 5072--5106.

\bibitem[{{Google}(2026)}]{google2026gemini4argon}
{Google}. 2026.
\newblock {Gemini 4 Argon}: Our next era of frontier intelligence.
\newblock
  \url{https://blog.google/innovation-and-ai/models-and-research/gemini-models/gemini-4-argon/}.

\bibitem[{Gopnik et~al.(2004)Gopnik, Glymour, Sobel, Schulz, Kushnir, and
  Danks}]{gopnik2004theory}
Alison Gopnik, Clark Glymour, David~M. Sobel, Laura~E. Schulz, Tamar Kushnir,
  and David Danks. 2004.
\newblock A theory of causal learning in children: Causal maps and {Bayes}
  nets.
\newblock \emph{Psychological Review}, 111(1):3--32.

\bibitem[{Harris and Koenig(2006)}]{harris2006trust}
Paul~L. Harris and Melissa~A. Koenig. 2006.
\newblock Trust in testimony: How children learn about science and religion.
\newblock \emph{Child Development}, 77(3):505--524.

\bibitem[{Ilievski et~al.(2021)Ilievski, Oltramari, Ma, Zhang, McGuinness, and
  Szekely}]{ilievski2021dimensions}
Filip Ilievski, Alessandro Oltramari, Kaixin Ma, Bin Zhang, Deborah~L.
  McGuinness, and Pedro Szekely. 2021.
\newblock Dimensions of commonsense knowledge.
\newblock \emph{arXiv preprint arXiv:2101.04640}.

\bibitem[{Jansen et~al.(2024)Jansen, C{\^o}t{\'e}, Khot, Bransom, Dalvi~Mishra,
  Majumder, Tafjord, and Clark}]{jansen2024discoveryworld}
Peter Jansen, Marc-Alexandre C{\^o}t{\'e}, Tushar Khot, Erin Bransom, Bhavana
  Dalvi~Mishra, Bodhisattwa~Prasad Majumder, Oyvind Tafjord, and Peter Clark.
  2024.
\newblock {DiscoveryWorld}: A virtual environment for developing and evaluating
  automated scientific discovery agents.
\newblock In \emph{Advances in Neural Information Processing Systems, Datasets
  and Benchmarks Track}.

\bibitem[{Jiang et~al.(2025)Jiang, Liu, Chen, Luo, Chen, Pan, Li, and
  Lin}]{jiang2025express}
Kaixuan Jiang, Yang Liu, Weixing Chen, Jingzhou Luo, Ziliang Chen, Ling Pan,
  Guanbin Li, and Liang Lin. 2025.
\newblock Beyond the destination: A novel benchmark for exploration-aware
  embodied question answering.
\newblock In \emph{Proceedings of the IEEE/CVF International Conference on
  Computer Vision}.

\bibitem[{Jin et~al.(2026)Jin, Liang, and Shen}]{jin2026stage}
Baosheng Jin, Yushen Liang, and Hua Shen. 2026.
\newblock {STAGE}: Diagnosing semantic transfer at grounded execution in
  embodied agents.
\newblock In \emph{Findings of the Association for Computational Linguistics:
  EMNLP 2026}, pages 24309--24331.

\bibitem[{Ju et~al.(2026)Ju, Sun, Wu, Zhang, Huo, Su, Gu, Cai, Liu, and
  Zhang}]{ju2026mineexplorer}
Tianjie Ju, Yueqing Sun, Zheng Wu, Wei Zhang, Yaqi Huo, Xi~Su, Qi~Gu, Xunliang
  Cai, Gongshen Liu, and Zhuosheng Zhang. 2026.
\newblock {MineExplorer}: Evaluating open-world exploration of {MLLM} agents in
  {Minecraft}.
\newblock \emph{arXiv preprint arXiv:2605.30931}.

\bibitem[{Li et~al.(2026)Li, Zheng, Song, Wong, and See}]{li2026ultradiscovery}
Weihan Li, Tianshi Zheng, Yangqiu Song, Ginny~Y. Wong, and Simon See. 2026.
\newblock {UltraDiscovery}: Abductive exploration in an interconnected,
  epistemically open universe.
\newblock \emph{arXiv preprint arXiv:2610.03092}.

\bibitem[{Liao et~al.(2026)Liao, Wang, Zhu, Du, Yan, and
  Chen}]{liao2026beliefmem}
Junfeng Liao, Qizhou Wang, Jianing Zhu, Bo~Du, Rui Yan, and Xiuying Chen. 2026.
\newblock Belief memory: Agent memory under partial observability.
\newblock \emph{arXiv preprint arXiv:2605.05583}.

\bibitem[{Lin et~al.(2026)Lin, Wang, Hong, Xie, Zeng, and
  Liu}]{lin2026asiworld}
Jianfeng Lin, Runze Wang, Haorong Hong, Yuxuan Xie, Zeno Zeng, and Zhongyuan
  Liu. 2026.
\newblock {ASIWorld}: Benchmarking autonomous agents in a persistent survival
  sandbox.
\newblock \emph{arXiv preprint arXiv:2610.14807}.

\bibitem[{Lynch and Sermanet(2021)}]{lynch2021language}
Corey Lynch and Pierre Sermanet. 2021.
\newblock Language conditioned imitation learning over unstructured data.
\newblock In \emph{Robotics: Science and Systems}.

\bibitem[{Magne et~al.(2026)Magne, Awadalla, Wang et~al.}]{magne2026nitrogen}
Lo{\"i}c Magne, Anas Awadalla, Guanzhi Wang, et~al. 2026.
\newblock {NitroGen}: An open foundation model for generalist gaming agents.
\newblock \emph{arXiv preprint arXiv:2601.02427}.

\bibitem[{Majumdar et~al.(2024)Majumdar, Ajay, Zhang
  et~al.}]{majumdar2024openeqa}
Arjun Majumdar, Anurag Ajay, Xiaohan Zhang, et~al. 2024.
\newblock {OpenEQA}: Embodied question answering in the era of foundation
  models.
\newblock In \emph{Proceedings of the IEEE/CVF Conference on Computer Vision
  and Pattern Recognition}.

\bibitem[{Meng et~al.(2026)Meng, Fan, Fan, Guo, Xie, and Han}]{meng2026ensimem}
Xuanyu Meng, Xing Fan, Xinyi Fan, Chenlei Guo, Yixuan Xie, and Jiawei Han.
  2026.
\newblock {EnSIMem}: Entity-structured indexing for long-term agent memory.
\newblock \emph{arXiv preprint arXiv:2609.27279}.

\bibitem[{{Mobius Digital}(2019)}]{outerwilds2019}
{Mobius Digital}. 2019.
\newblock Outer wilds.
\newblock Video game, published by Annapurna Interactive.

\bibitem[{{OpenAI}(2026)}]{openai2026gpt61sol}
{OpenAI}. 2026.
\newblock Introducing {GPT-6.1 Sol}.
\newblock \url{https://openai.com/index/introducing-gpt-6-1-sol/}.

\bibitem[{{Outer Wilds Wiki contributors}(2026)}]{outerwildswiki}
{Outer Wilds Wiki contributors}. 2026.
\newblock Outer wilds wiki.
\newblock \url{https://outerwilds.fandom.com}.
\newblock Accessed November 2026.

\bibitem[{Ouyang et~al.(2026)Ouyang, Hu, Lin, Ng, and
  Shou}]{ouyang2026gameworld}
Mingyu Ouyang, Siyuan Hu, Kevin~Qinghong Lin, Hwee~Tou Ng, and Mike~Zheng Shou.
  2026.
\newblock {GameWorld}: Towards standardized and verifiable evaluation of
  multimodal game agents.
\newblock \emph{arXiv preprint arXiv:2604.07429}.

\bibitem[{Panickssery et~al.(2024)Panickssery, Bowman, and
  Feng}]{panickssery2024llm}
Arjun Panickssery, Samuel~R. Bowman, and Shi Feng. 2024.
\newblock {LLM} evaluators recognize and favor their own generations.
\newblock In \emph{Advances in Neural Information Processing Systems}.

\bibitem[{Park et~al.(2023)Park, O'Brien, Cai, Morris, Liang, and
  Bernstein}]{park2023generative}
Joon~Sung Park, Joseph~C. O'Brien, Carrie~J. Cai, Meredith~Ringel Morris, Percy
  Liang, and Michael~S. Bernstein. 2023.
\newblock Generative agents: Interactive simulacra of human behavior.
\newblock In \emph{Proceedings of the 36th Annual ACM Symposium on User
  Interface Software and Technology (UIST)}.

\bibitem[{Perez et~al.(2018)Perez, Strub, de~Vries, Dumoulin, and
  Courville}]{perez2018film}
Ethan Perez, Florian Strub, Harm de~Vries, Vincent Dumoulin, and Aaron
  Courville. 2018.
\newblock {FiLM}: Visual reasoning with a general conditioning layer.
\newblock In \emph{Proceedings of the AAAI Conference on Artificial
  Intelligence}.

\bibitem[{Raffel et~al.(2020)Raffel, Shazeer, Roberts, Lee, Narang, Matena,
  Zhou, Li, and Liu}]{raffel2020t5}
Colin Raffel, Noam Shazeer, Adam Roberts, Katherine Lee, Sharan Narang, Michael
  Matena, Yanqi Zhou, Wei Li, and Peter~J. Liu. 2020.
\newblock Exploring the limits of transfer learning with a unified text-to-text
  transformer.
\newblock \emph{Journal of Machine Learning Research}, 21(140):1--67.

\bibitem[{Rashkin et~al.(2023)Rashkin, Nikolaev, Lamm, Aroyo, Collins, Das,
  Petrov, Tomar, Turc, and Reitter}]{rashkin2023ais}
Hannah Rashkin, Vitaly Nikolaev, Matthew Lamm, Lora Aroyo, Michael Collins,
  Dipanjan Das, Slav Petrov, Gaurav~Singh Tomar, Iulia Turc, and David Reitter.
  2023.
\newblock Measuring attribution in natural language generation models.
\newblock \emph{Computational Linguistics}, 49(4):777--840.

\bibitem[{Rasmussen et~al.(2025)Rasmussen, Paliychuk, Beauvais, Ryan, and
  Chalef}]{rasmussen2025zep}
Preston Rasmussen, Pavlo Paliychuk, Travis Beauvais, Jack Ryan, and Daniel
  Chalef. 2025.
\newblock {Zep}: A temporal knowledge graph architecture for agent memory.
\newblock \emph{arXiv preprint arXiv:2501.13956}.

\bibitem[{Shah et~al.(2025)Shah, Chen, Godbole, Mora, Seshia, and
  Levine}]{shah2025liten}
Ameesh Shah, William Chen, Adwait Godbole, Federico Mora, Sanjit~A. Seshia, and
  Sergey Levine. 2025.
\newblock Learning affordances at inference-time for vision-language-action
  models.
\newblock \emph{arXiv preprint arXiv:2510.19752}.

\bibitem[{Shelton et~al.(2025)Shelton, Wingerd, Rijal, Garg, Gutic, Barnes, and
  Finegan-Dollak}]{shelton2025pqb}
Miles Shelton, Nate Wingerd, Kritim~K. Rijal, Ayush Garg, Adelina Gutic, Brett
  Barnes, and Catherine Finegan-Dollak. 2025.
\newblock Grounded, or a good guesser? a per-question balanced dataset to
  separate blind from grounded models for embodied question answering.
\newblock In \emph{Proceedings of the 63rd Annual Meeting of the Association
  for Computational Linguistics (Volume 2: Short Papers)}, pages 124--135.

\bibitem[{Shinn et~al.(2023)Shinn, Cassano, Gopinath, Narasimhan, and
  Yao}]{shinn2023reflexion}
Noah Shinn, Federico Cassano, Ashwin Gopinath, Karthik Narasimhan, and Shunyu
  Yao. 2023.
\newblock Reflexion: Language agents with verbal reinforcement learning.
\newblock In \emph{Advances in Neural Information Processing Systems}.

\bibitem[{Tulving(1972)}]{tulving1972episodic}
Endel Tulving. 1972.
\newblock Episodic and semantic memory.
\newblock In Endel Tulving and Wayne Donaldson, editors, \emph{Organization of
  Memory}, pages 381--403. Academic Press.

\bibitem[{Verga et~al.(2024)Verga, Hofstatter, Althammer, Su, Piktus,
  Arkhangorodsky, Xu, White, and Lewis}]{verga2024replacing}
Pat Verga, Sebastian Hofstatter, Sophia Althammer, Yixuan Su, Aleksandra
  Piktus, Arkady Arkhangorodsky, Minjie Xu, Naomi White, and Patrick Lewis.
  2024.
\newblock Replacing judges with juries: Evaluating {LLM} generations with a
  panel of diverse models.
\newblock \emph{arXiv preprint arXiv:2404.18796}.

\bibitem[{Wang et~al.(2023)Wang, Xie, Jiang, Mandlekar, Xiao, Zhu, Fan, and
  Anandkumar}]{wang2023voyager}
Guanzhi Wang, Yuqi Xie, Yunfan Jiang, Ajay Mandlekar, Chaowei Xiao, Yuke Zhu,
  Linxi Fan, and Anima Anandkumar. 2023.
\newblock Voyager: An open-ended embodied agent with large language models.
\newblock \emph{arXiv preprint arXiv:2305.16291}.

\bibitem[{Wang et~al.(2026)Wang, Lai, Liu et~al.}]{wang2026adaexplorebench}
Rui Wang, Zishen Lai, Jiayu Liu, et~al. 2026.
\newblock {AdaExploreBench}: Benchmarking {LLM} agents for proactive
  exploration and test-time adaptation in evolving environments.
\newblock \emph{arXiv preprint arXiv:2610.25950}.

\bibitem[{Wei et~al.(2025)Wei, Sachdeva, Coleman et~al.}]{wei2025evomemory}
Tianxin Wei, Noveen Sachdeva, Benjamin Coleman, et~al. 2025.
\newblock {Evo-Memory}: Benchmarking {LLM} agent test-time learning with
  self-evolving memory.
\newblock \emph{arXiv preprint arXiv:2511.20857}.

\bibitem[{Wu et~al.(2026)Wu, Zhou, Zhao, Cao, Pu, and Yang}]{wu2026abstaineqa}
Tao Wu, Chuhao Zhou, Guangyu Zhao, Haozhi Cao, Yewen Pu, and Jianfei Yang.
  2026.
\newblock When robots should say ``{I} don't know'': Benchmarking abstention in
  embodied question answering.
\newblock In \emph{Proceedings of the IEEE/CVF Conference on Computer Vision
  and Pattern Recognition}.

\bibitem[{Xiao et~al.(2021)Xiao, Shang, Yao, and Chua}]{xiao2021nextqa}
Junbin Xiao, Xindi Shang, Angela Yao, and Tat-Seng Chua. 2021.
\newblock {NExT-QA}: Next phase of question-answering to explaining temporal
  actions.
\newblock In \emph{Proceedings of the IEEE/CVF Conference on Computer Vision
  and Pattern Recognition}.

\bibitem[{Xiao et~al.(2024)Xiao, Yao, Li, and Chua}]{xiao2024nextgqa}
Junbin Xiao, Angela Yao, Yicong Li, and Tat-Seng Chua. 2024.
\newblock Can {I} trust your answer? visually grounded video question
  answering.
\newblock In \emph{Proceedings of the IEEE/CVF Conference on Computer Vision
  and Pattern Recognition}.

\bibitem[{Xiao et~al.(2023)Xiao, Chan, Sermanet, Wahid, Brohan, Hausman,
  Levine, and Tompson}]{xiao2023dial}
Ted Xiao, Harris Chan, Pierre Sermanet, Ayzaan Wahid, Anthony Brohan, Karol
  Hausman, Sergey Levine, and Jonathan Tompson. 2023.
\newblock Robotic skill acquisition via instruction augmentation with
  vision-language models.
\newblock In \emph{Robotics: Science and Systems}.

\bibitem[{Zhang et~al.(2025)Zhang, Griffiths, Narasimhan, and
  Press}]{zhang2025videogamebench}
Alex~L. Zhang, Thomas~L. Griffiths, Karthik~R. Narasimhan, and Ofir Press.
  2025.
\newblock {VideoGameBench}: Can vision-language models complete popular video
  games?
\newblock \emph{arXiv preprint arXiv:2505.18134}.

\bibitem[{Zhang et~al.(2023)Zhang, Pertsch, Zhang, and Lim}]{zhang2023sprint}
Jesse Zhang, Karl Pertsch, Jiahui Zhang, and Joseph~J. Lim. 2023.
\newblock {SPRINT}: Scalable policy pre-training via language instruction
  relabeling.
\newblock \emph{arXiv preprint arXiv:2306.11886}.

\bibitem[{Zhang et~al.(2026{\natexlab{a}})Zhang, Xiang, Gao
  et~al.}]{zhang2026explorationbench}
Ming Zhang, Zhenghao Xiang, Peizhong Gao, et~al. 2026{\natexlab{a}}.
\newblock {ExplorationBench}: Measuring {AI} systems' exploration in verifiable
  alien worlds.
\newblock \emph{arXiv preprint arXiv:2609.30199}.

\bibitem[{Zhang et~al.(2026{\natexlab{b}})Zhang, Wen, Zhang, Wang, Xie,
  Khashabi, and Shu}]{agentodyssey2026}
Zheyuan Zhang, Zehao Wen, Alvin Zhang, Andrew Wang, Jianwen Xie, Daniel
  Khashabi, and Tianmin Shu. 2026{\natexlab{b}}.
\newblock {AgentOdyssey}: Open-ended long-horizon text game generation for
  test-time continual learning agents.
\newblock \emph{arXiv preprint arXiv:2606.24893}.

\bibitem[{Zheng et~al.(2023)Zheng, Chiang, Sheng, Zhuang, Wu, Zhuang, Lin, Li,
  Li, Xing, Zhang, Gonzalez, and Stoica}]{zheng2023judging}
Lianmin Zheng, Wei-Lin Chiang, Ying Sheng, Siyuan Zhuang, Zhanghao Wu, Yonghao
  Zhuang, Zi~Lin, Zhuohan Li, Dacheng Li, Eric~P. Xing, Hao Zhang, Joseph~E.
  Gonzalez, and Ion Stoica. 2023.
\newblock Judging {LLM}-as-a-judge with {MT-Bench} and chatbot arena.
\newblock In \emph{Advances in Neural Information Processing Systems, Datasets
  and Benchmarks Track}.

\end{thebibliography}

\end{document}